\documentclass[letterpaper, 10 pt, conference]{ieeeconf}  

\IEEEoverridecommandlockouts                              

\usepackage{formatting} 
\usepackage{multirow} 

\usepackage{booktabs}
\usepackage{array}
\usepackage{pifont}
\usepackage{makecell}

\newcolumntype{P}[1]{>{\centering\arraybackslash}p{#1}}
\newcolumntype{M}[1]{>{\centering\arraybackslash}m{#1}}
\newcommand{\cmark}{\textcolor{black}{\ding{51}}}

\usepackage{hyperref}
\usepackage{comment}

\title{\LARGE \bf
Dynamic Layer Detection of Thin Materials using DenseTact Optical Tactile Sensors}

\author{Ankush Kundan Dhawan$^{\dagger}$, Camille Chungyoun$^{\dagger}$, Karina Ting$^{\dagger}$, and Monroe Kennedy III 
\thanks{Authors are members of the ARMLab in the Mechanical Engineering Department, Stanford University, Stanford, CA 94305, USA. 
{\tt\small \{ankushd, camillec , khting, monroek\}@stanford.edu.} 
This work is supported by the National Science Foundation under Grants 2142773 and 2220867. Project page is available at \url{https://armlabstanford.github.io/dynamic-cloth-detection}.  
}  
\thanks{$^\dagger$ authors contributed equally to this work.}}
\begin{document}

\maketitle
\thispagestyle{empty}
\pagestyle{empty}

\begin{abstract}

Manipulation of thin materials is critical for many everyday tasks and remains a significant challenge for robots. While existing research has made strides in tasks like material smoothing and folding, many studies struggle with common failure modes (crumpled corners/edges, incorrect grasp configurations) that a preliminary step of layer detection could solve. We present a novel method for classifying the number of grasped material layers using a custom gripper equipped with DenseTact 2.0 optical tactile sensors. After grasping, the gripper performs an anthropomorphic rubbing motion while collecting optical flow, 6-axis wrench, and joint state data. Using this data in a transformer-based network achieves a test accuracy of 98.21\% in classifying the number of grasped cloth layers, and 81.25\% accuracy in classifying layers of grasped paper, showing the effectiveness of our dynamic rubbing method. Evaluating different inputs and model architectures highlights the usefulness of tactile sensor information and a transformer model for this task. A comprehensive dataset of 568 labeled trials (368 for cloth and 200 for paper) was collected and made open-source along with this paper. 
\end{abstract}

\section{Introduction}
The ability of robots to manipulate common objects found in the home is essential for their utility and ubiquity \cite{assistive_robots_review}. Thin, deformable materials such as cloth are still a challenge for robots for tasks such as folding laundry, assistive dressing, making the bed, and sewing \cite{unfolding_literature, grippers_soft, wang2023policy}. Similarly, paper is another thin, deformable material that is used in everyday tasks, such as turning pages and organizing documents.

Deformable objects, such as cloth and paper, may exhibit complex dynamics, high degrees of freedom, and self-occlusion, posing significant obstacles for traditional manipulation methods \cite{zhu2022challenges}. Conventional approaches often rely on electrical signal processing or mechanical system identification \cite{fu2024mobile}, which may struggle to adapt to the nuanced and variable nature of these material interactions. As a result of these challenging properties, these materials often become folded or bunched up during manipulation \cite{sunil2022visuotactile}, leading to poor execution of tasks. 

\begin{figure}[t]
    \vskip 0.04in
    \begin{center}
    \includegraphics[width=\columnwidth]{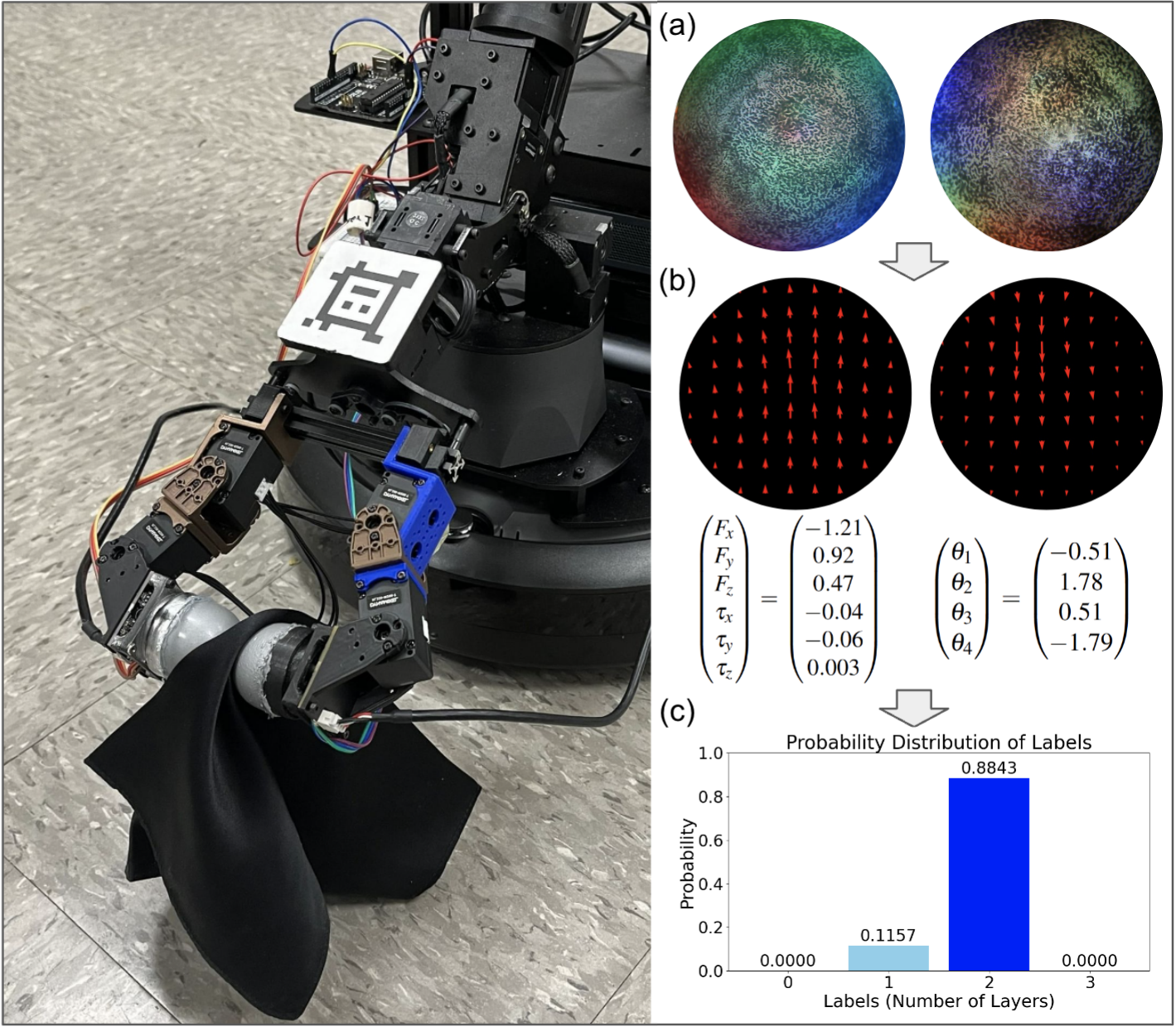}
    \caption{\textbf{Overview of Layer Classification} Our gripper rubs a material between its fingers for layer classification. As shown on the left, the gripper can be mounted directly on a LoCoBot for more complex tasks. The right shows (a) the DenseTact RGB images, (b) the model inputs of optical flow, wrench, and joint states, and (c) the classification results.}
    \label{fig:main}
    \end{center}
    \vspace{-20 pt}
\end{figure}


Inspired by human dexterous motion, we present a method for identifying the number of layers of cloth or paper grasped by a robotic gripper using a dynamic rubbing motion. We selected cloth and paper as materials as both are often used in everyday tasks. Our method uses a custom gripper equipped with optical tactile sensors. To classify the number of layers, the gripper performs an anthropomorphic rubbing motion while recording optical flow, 6-axis wrench, and joint state data. This data is used in a transformer-based network to identify the number of grasped material layers.

The main contributions of this paper are:
\begin{enumerate}
    \item A compact, 4 DOF gripper equipped with DenseTact 2.0 sensors, capable of performing a rubbing motion between its fingers (Section \ref{subsec:gripper_design}).
    \item A dataset for layer classification based on tactile sensor output. Included classes are 0, 1, 2, and 3 layers of cloth and paper (Section \ref{section: data_collection}).
    \item A transformer-based network 
    that classifies the number of cloth or paper layers using optical flow, wrench, and joint state data taken during the rubbing motion (Section \ref{section: classifier_network}). The network can run in real time to classify layers based on the most recent data at 3 Hz.
\end{enumerate}

The paper is structured as follows. Section \ref{sec:related} covers related works. Section \ref{sec:hardware} presents the gripper design and DenseTact 2.0 sensors. Section \ref{sec:methods} describes the input data modalities and network architecture. Section \ref{sec:experiments} outlines the experiments conducted to formulate and validate our model. Section \ref{sec:conclusion} discusses conclusions and future work. 

\begin{figure}[htbp]
    \centering
    \includegraphics[height=6 cm]
    {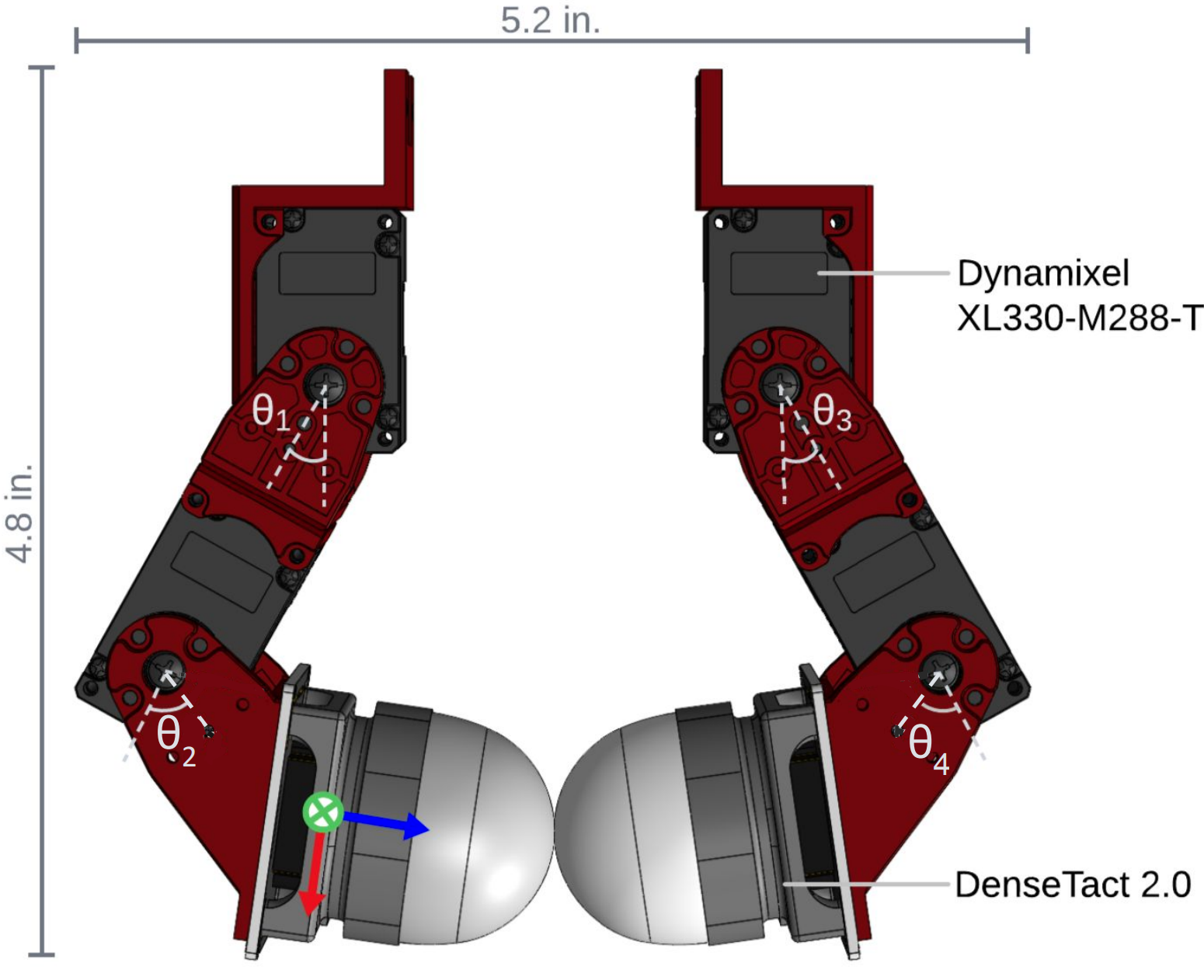}
    \caption{\textbf{Gripper Design}: Model of custom gripper with DenseTact sensors. Shown in the grasping position with labeled joints and DenseTact axes for the 6-axis wrench.}
    \label{fig:gripper_cad}
    \vspace{-10 pt}
\end{figure}


\section{Related Works} \label{sec:related}

Much of the existing literature on thin material manipulation focuses on smoothing and folding tasks \cite{cloth_smoothing, fu2024mobile}. While the results are promising, these works show room for improvement with common failure modes. In thin material smoothing works, many end-results still have folded or crumpled corners and edges  \cite{wu2020manipulatedeformableobjects, seita2020smoothing}. Similarly, in thin material folding works, the achieved configuration differs from the target when more than one layer was mistakenly grasped  \cite{ganapathi2021simfold, weng2021fabricflownet}.

These failure modes could be avoided using layer detection. Tactile information is especially useful for this task, yet utilizing tactile sensors for the manipulation of thin, deformable materials is less explored. In \cite{yuan2018activeclothingmaterialperception}, the effectiveness of using optical sensors to identify the physical properties of a thin material is shown. By squeezing various materials using GelSight sensors, they are able to recognize textural differences. For material layer detection, \cite{tirumala2022learning} approaches the task using magnetometer-based tactile ReSkin sensors. While their method consistently classifies 0 and 1 layer of the material, it struggles more with 2 and 3 layers, with accuracies of 86.6\% and 47.8\%, respectively. In addition, their method using tactile sensing from a static grasp would not be effective for smooth-textured or thin materials. In \cite{jiang2024rotipbot}, a rotating vision-based tactile sensor is used to grasp and count layers of thin, flexible materials. However, their method relies on a sequential layer-counting process, which is inherently time-consuming and inefficient in tasks requiring rapid manipulation or those involving thinner cloths. Previous work has not explored using anthropomorphic rubbing with optical tactile sensors for thin material layer classification.

For our method, we leverage friction as a tool for layer detection. Using optical tactile sensors in a rubbing motion provides dynamic information that cannot be obtained from a static grasp. This approach is anthropomorphically motivated, as humans might perform a similar rubbing motion to accomplish this task. Additionally, our method is able to efficiently detect the number of layers in a single dynamic grasp, eliminating the need for sequential counting. We use DenseTact 2.0 optical tactile sensors for their high resolution and rich feature visualization \cite{do2023densetact}. Building upon the principles outlined in \cite{do2023densetact} and \cite{do2022densetact}, we utilize RGB videos of the sensor gel deformations to record wrench estimation and optical flow data.

\begin{figure}[htbp]
    \centering
    \includegraphics[height=7 cm]
    {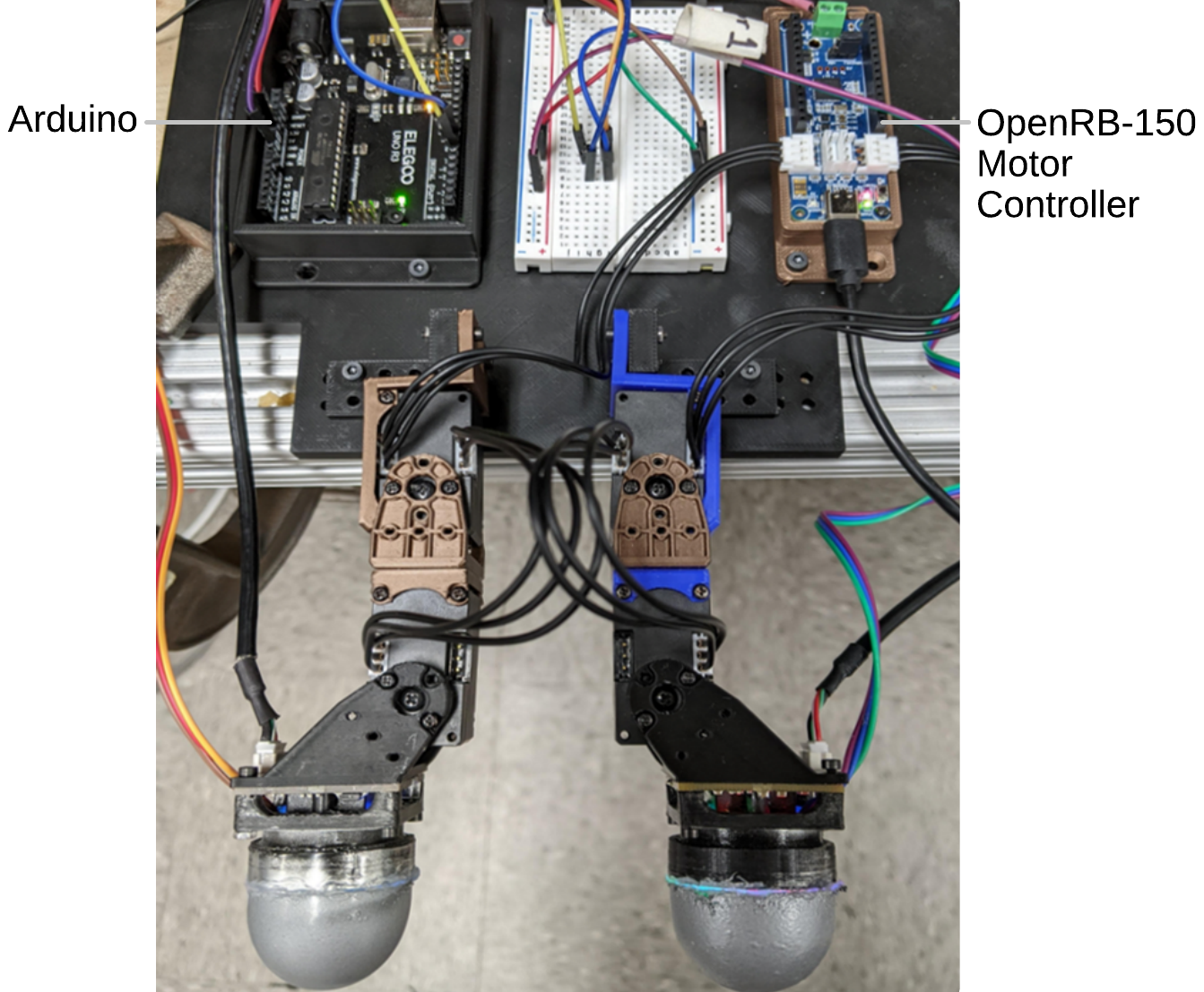}
    \caption{\textbf{Gripper System Setup}: The hardware setup for data collection. The gripper starts in the configuration shown.}
    \label{fig:gripper}
    \vspace{-10 pt}
\end{figure}

\section{Hardware Setup}\label{sec:hardware}
The hardware developed for this research project is a two-finger robotic gripper with DenseTact 2.0 sensors as fingertips. The gripper components communicate via ROS2 to perform robotic layer classification.

\subsection{Two-Finger Gripper Design} \label{subsec:gripper_design}

A custom two-finger, 4 DOF gripper was designed, as shown in Fig. \ref{fig:gripper_cad}. Each finger uses two Dynamixel XL330-M288-T motors as revolute joints, chosen for their lightweight and compact nature. To emulate finger pads, a DenseTact 2.0 sensor is attached at the tip of each finger. An OpenRB-150 Arduino-compatible embedded controller is used for motor control, and an Arduino board is used to run the DenseTact sensors. Custom 3D printed mounts integrate the motors and DenseTact sensors together. A single computer ran the central code for controlling the robotic system that also provided power to the motor control board and the DenseTact cameras and LEDs.

\begin{figure*}[htbp]
    \centering
    \includegraphics[width=15cm]{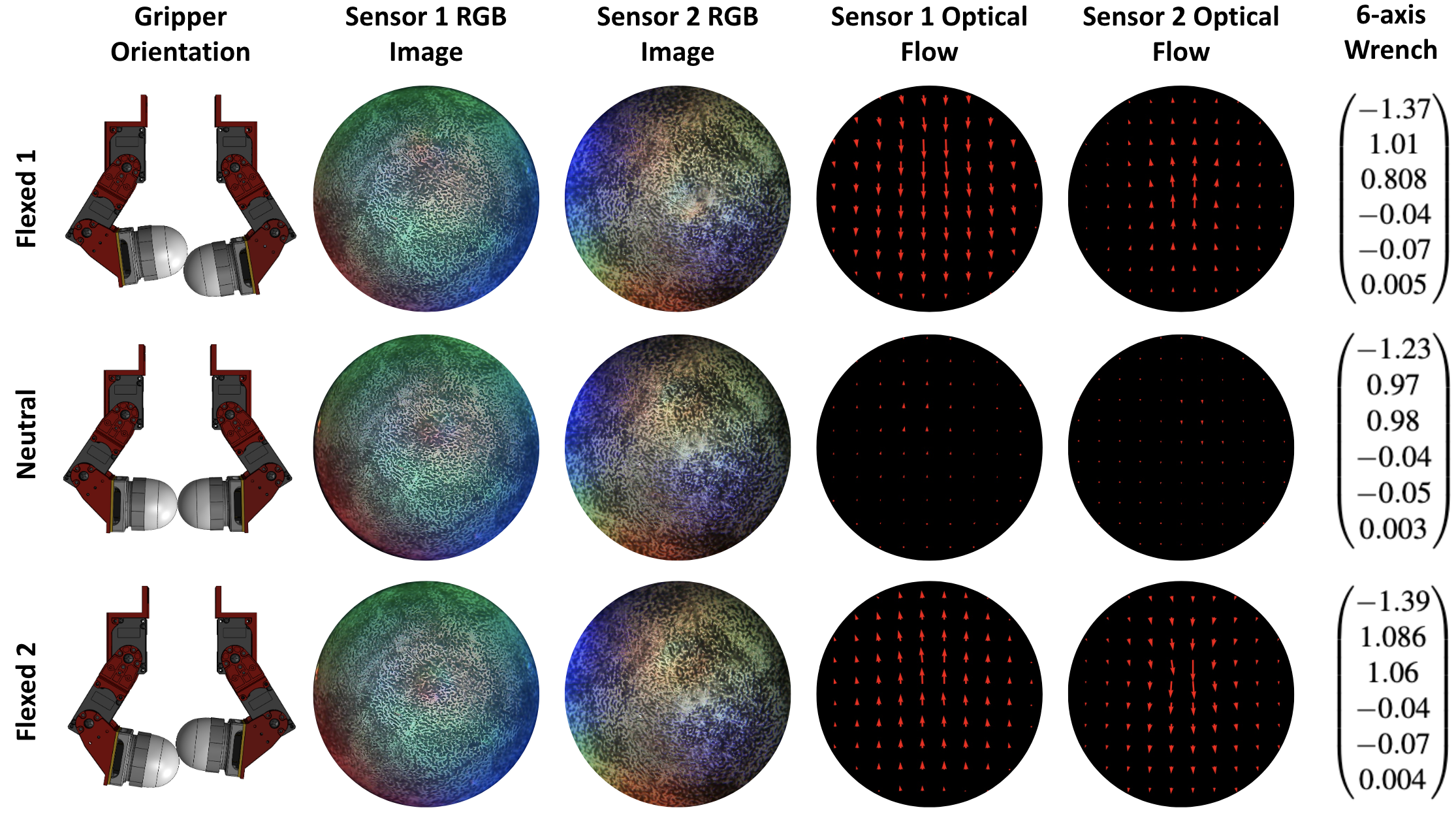}
    \caption{\textbf{Optical Flow and Wrench Data based on Gripper Orientation}: The DenseTact gel pattern provides a camera feed with rich trackable features. The optical flow between consecutive frames provides the magnitude and direction of feature movement during the rubbing motion. The calibrated DenseTact outputs net wrench data, while the joint states are recorded across the entire rubbing motion.}
    \label{fig:pattern}
    \vspace{-10 pt}
\end{figure*}

\subsection{DenseTact 2.0}

The DenseTact sensor acts as a fingertip, providing a deformable medium with tactile sensing capabilities. The sensor includes an integrated camera, RGB LEDs, and a gel surface with a randomized pattern (shown in Fig. \ref{fig:pattern}), allowing for a feature-rich camera feed of gel deformation during manipulation tasks \cite{do2023densetact}. Calibrated DenseTact sensors additionally provide 6-axis wrench estimation. DenseTact 2.0 was chosen over other designs such as DenseTact 1.0 \cite{do2022densetact} and DenseTact Mini \cite{do2023densetactmini} because of its compact design, yet high resolution and hemispherical shape. This hemispherical shape, along with the deformable and high-friction nature of the gel surface, make DenseTact 2.0 especially effective in grasping and manipulating thin materials.




\section{Methods} \label{sec:methods}
\subsection{Optical Flow} \label{section: optical_flow}
The randomized pattern on the DenseTact surface allows for a feature-rich image even when the objects being manipulated are relatively smooth. This makes optical flow an effective method of quantifying movement at the gel's surface during manipulation tasks, such as the rubbing motion used to classify the number of grasped layers. The Farnebäck dense optical flow method \cite{farneback_dense_optical_flow} was used to estimate the motion between two frames. This method outputs a matrix, where each element (x, y) is a 2D motion vector (u, v) that indicates the displacement of the points between the consecutive frames. Each vector (u, v) at every position in the matrix describes how much the pixel at (x, y) has moved in the x-direction (u) and the y-direction (v). To visualize this, optical flow can be represented in a quiver plot, where the magnitude and direction of the arrows at each pixel indicate the displacement. 

\subsection{6-Axis Wrench} \label{section: wrench}
Net force and torque data are also potentially informative of the interactions between the DenseTact sensors and the thin material. Calibrating DenseTact sensors following the method in \cite{do2023densetact} allows for real-time 6-axis wrench estimation ($F_x, F_y, F_z, \tau_x, \tau_y, \tau_z$) for a given RGB image frame. Once calibrated, feeding an RGB image from the DenseTact in a forward pass through the calibrated model returns the wrench estimation. This data was collected per image frame and recorded for every trial in the dataset.

\subsection{Joint States}
The angles of the 4 motor joints of the custom gripper were recorded and used as an input to the model, measured as the motor angle. Since the motion for every trial was relatively similar, the model could associate the other inputs with specific joint states that are affected by the friction involved with the dynamic grasp.

\subsection{Dataset Collection} \label{section: data_collection}
A custom dataset was collected to train, validate, and test the classifier. The cloth used was a black silk handkerchief with a single-layer thickness of 0.16 mm (see Fig \ref{fig:main}). The paper was standard printer paper. These materials were chosen for their few visual features and low thickness, providing a good baseline for generalizability to other thin textiles. The dataset is comprised of RGB videos of gel deformations and 6-axis wrench estimation captured by DenseTact, as well as the gripper's joint states. All data was collected in real-time while the gripper performed a 16-second long rubbing motion. As described in Section \ref{sec:data_params}, the period of the sine motion was 4 seconds, meaning there were 4 cycles over the full trial. The dataset is labeled with the ground truth number of grasped layers. A total of 92 trials for each class of cloth and 50 trials for each class of paper was collected, giving a total of 568 examples across the entire dataset. Each video trial was recorded at 10 Hz and pre-processed before being input to the classifier network.

\begin{figure*}[ht]
    \centering
    \includegraphics[width=17 cm]{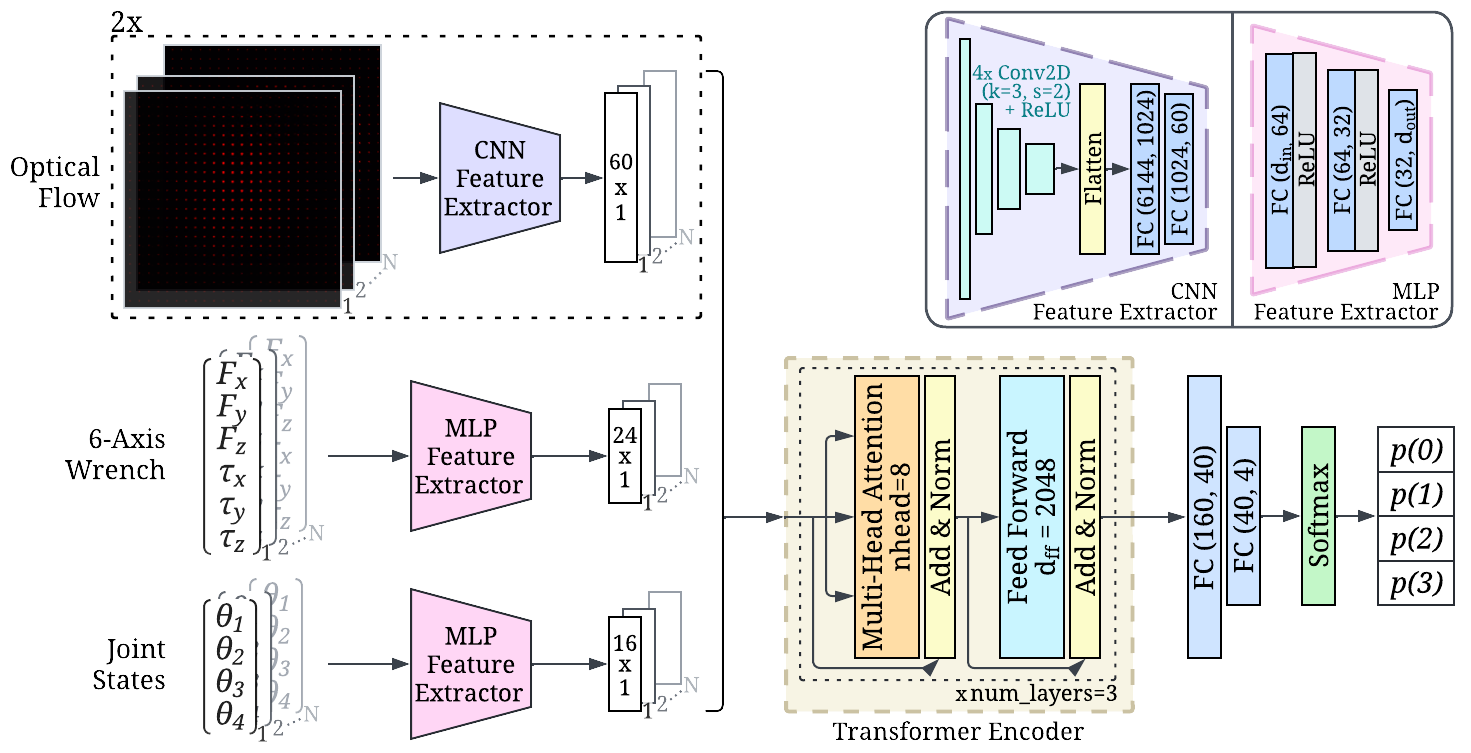}
    \caption{\textbf{Network Architecture for Layer Classifier}: The model inputs are N-length time sequences of optical flow, 6-axis wrench, and joint state data (N=200). Extracted features from each input are concatenated and fed into a transformer encoder \cite{vaswani2017attention}, followed by fully-connected layers and a softmax function. The output is the probability distribution across the 4 classes (0, 1, 2 and 3 layers). The parameter values for the transformer encoder are $d_{model}=160$, $nhead=8$, $num\_layers=3$, and $d_{ff}=2048$, where $d_{model}$ is the dimension of the input and output embedding, $nhead$ is the number of attention heads, $num\_layers$ is the number of encoder layers, and $d_{ff}$ is the feed-forward network's inner layer dimension.}
    \label{fig: architecture}
    \vspace{-10 pt}
\end{figure*}

\subsection{Data Pre-Processing} \label{section: pre-process}
In order to provide meaningful data to the network, the optical flow data frame rate was modified during pre-processing. After experimentation with sub-sampling and visual inspection of the resulting vectors, the camera's frame rate of 10 Hz was found to be a high enough frequency where the optical flows accurately represented the rubbing motion without aliasing, and a low enough frequency where the differences between consecutive frames were sufficiently large to be informative. To the same effect, the optical flow measurements were average pooled by 8 (a window of 64 pixels). After pooling, the dimensions of the optical flow inputs were $96\times128\times2$, indicating the image height, image width, and the number of channels (magnitude and direction). For consistency, all other inputs were also sub-sampled to a 10 Hz frequency. 

\subsection{Classification Model} \label{section: classifier_network}
The network architecture for this classification task is shown in Fig. \ref{fig: architecture}, where the inputs are N-length sequences of optical flow, wrench, and joint state data. The model consists of a feature extraction stage and a classification stage.


\subsubsection{Feature Extraction}
Each input is passed through a feature extractor to capture notable patterns that differentiate the classes. CNN and ResNet backbones were tested for the optical flow feature extractor. The CNN feature extractor performed better due to the dataset's small size, as shown in Table \ref{tab:classifier_comparison}. Hence, to improve computational efficiency, a CNN structure was chosen over the more complex ResNet structure. This is described further in Section \ref{section: ablation}.
The wrench and joint state feature extractors use a simple MLP with fully connected layers and ReLU functions. Since the input dimensions per frame are small ($d_{in}=6$ and $d_{in}=4$, respectively) compared to the optical flow feature vector, these MLPs increase the dimensions to $d_{out}=24$ and $d_{out}=16$, respectively. 

\subsubsection{Classification}
The extracted input features are concatenated and fed into a transformer encoder \cite{vaswani2017attention}. The transformer encoder was chosen for this task because the inputs are time sequences of data, and the transformer is able to integrate spatiotemporal and tactile features into a coherent representation of the rubbing motion using attention mechanisms. This allows the network to understand the feature time-dependence in the data collected during a rubbing motion.
The integrated features from the transformer are reduced to 4 dimensions using fully-connected layers. Finally, a softmax function is used, where the highest probability determines the predicted number of grasped layers (0, 1, 2, or 3 layers of the thin material).

\subsubsection{Model Training} 
We ran our model on the NVIDIA GeForce RTX 4080 GPU. We used a 70\%-15\%-15\% split to divide our collected data into training, validation, and test sets. To train our classifier model, we use the cross entropy loss function and the Adam optimizer with a learning rate of $5 \times 10^{-5}$. Optimizer state resets were performed every 10 epochs to improve training stability \cite{asadi2024resetting}. The small gaps between training and validation accuracy suggest generalization without overfitting and adequate training data.




\begin{table*}[ht]
    \def\arraystretch{1.2}
    \centering
    \resizebox{\textwidth}{!}{
    \begin{tabular}
    { @{\extracolsep{5pt}} l P{1.2cm} P{1.2cm} P{1.2cm} P{1cm} P{1cm} P{1cm} P{1.6cm} }

    \toprule
    & \multicolumn{7}{c}{Inputs} \\
    \cmidrule{3-7}
    \multirow{2}{*}{Backbone} & \multirow{2}{*}{Classifier} & S1 Flow Mags. & S2 Flow Mags. & S1 Flow Angles & S1 6-Axis Wrench & F1, F2 Joint States & Test Accuracy (\%) \\
    \midrule
    - & \text{Naive CNN}    & \cmark &        &        &        &        & 25.00 \\
    \Xhline{0.1pt}
    - & \text{Naive CNN*}   & \cmark &        &        &        &        & 89.29 \\
    \Xhline{0.1pt}
    -  & \text{ResNet-18} & \cmark  &        &        &        &     & 91.07 \\
    \Xhline{0.1pt}
    ResNet-18  & Transformer & \cmark  &        &        &        &        & 92.86 \\
    \Xhline{0.1pt}
    ResNet-18  & Transformer & \cmark  & \cmark &        & \cmark &        & 92.86 \\
    \Xhline{0.1pt}
    ResNet-18  & Transformer & \cmark  & \cmark &        &        & \cmark & 83.93 \\
    \Xhline{0.1pt} 
    ResNet-18  & Transformer &  \cmark & \cmark &        & \cmark & \cmark & 91.07 \\
    \Xhline{0.1pt}  
    CNN & Transformer  & \cmark &        &        &        &        & 25.00 \\
    \Xhline{0.1pt}
    CNN* & Transformer & \cmark &        &        &        &        & 96.43 \\
    \Xhline{0.1pt}
    CNN & Transformer  & \cmark &        & \cmark &        &        & 91.07 \\
    \Xhline{0.1pt}
    CNN & Transformer  & \cmark &        &        & \cmark &        & 92.86 \\
    \Xhline{0.1pt}
    CNN & Transformer  & \cmark &        &        & \cmark & \cmark & 92.86 \\
    \Xhline{0.1pt}
    CNN & Transformer  & \cmark & \cmark &        &        &        & 25.00 \\
    \Xhline{0.1pt}
    CNN* & Transformer & \cmark & \cmark &        &        &        & 94.64 \\
    \Xhline{0.1pt}
    CNN & Transformer  & \cmark & \cmark &        & \cmark &        & 96.43 \\
    \Xhline{0.1pt}
    CNN* & Transformer & \cmark & \cmark &        & \cmark &        & 91.07 \\
    \Xhline{0.1pt}
    CNN*$^{(1)}$& Transformer & \cmark & \cmark &        &        & \cmark & 96.43 \\
    \Xhline{0.1pt}
    CNN$^{(2)}$ & Transformer  & \cmark & \cmark &        & \cmark & \cmark & \textbf{98.21} \\
    \Xhline{0.1pt}
    CNN*$^{(3)}$ & Transformer & \cmark & \cmark &        & \cmark & \cmark & 92.86 \\
    \bottomrule
    \end{tabular}
    }

    \caption{\textbf{Ablation Study Results}: Comparison of classifier models with different architectures and inputs, each with a total of 56 test trials. All data used for these models were cloth trials. * indicates modified architecture of the CNN feature extractor, specifically batchnorm layers and one dropout layer. S1 = sensor 1; S2 = sensor 2; F1 = finger 1; F2 = finger 2. The backbone column refers to the optical flow feature extractor backbone. Superscripts (1), (2), (3) note the model architecture used for testing with paper. Consistency of performance across ablations indicates sufficient dataset size for training.}
    \label{tab:classifier_comparison}
    \vspace{-10pt}
\end{table*}

\begin{table}[htbp]
    \centering
    \normalsize
    \begin{tabular}{lc}
    \toprule
    \textbf{Model} & \textbf{Test Accuracy (\%)} \\
    \midrule
    Model 1 & 71.88 \\
    Model 2 & 81.25 \\
    Model 3 & 75.00 \\
    \bottomrule
    \end{tabular}
    \caption{\textbf{Comparison of Model Accuracies when tested on Paper Trials}. Model architectures are described in Table \ref{tab:classifier_comparison}, where the Transformer classifier with CNN backbone, with the optical flow, joint, and wrench feature extractors, performed the best at 81.25\% accuracy on unseen data.}
    \label{tab:model_comparison}
    \vspace{-15pt}
\end{table}

\section{Experiments} \label{sec:experiments}
\subsection{Determining Data Collection Parameters} \label{sec:data_params}
Parameters for data collection were tuned to ensure that meaningful data was input to the classifier. An x-direction force threshold was used to determine the right amount of contact between the DenseTact sensors, noted by a contact patch in the video stream where the sensors touched. Too large of a contact patch would prevent slipping between the fingers, saturating the optical flow data and muting informative friction data, while too small of a contact patch would cause a failed grasp. The force threshold was tuned to accommodate both factors. The tuned threshold value was tight enough for the DenseTact sensors to touch when there was no grasped material between them, and did not create too large of a contact patch when there were 3 layers of thin material between the DenseTact sensors. This increased the interpretability of the data for the model.

The period of the sine wave performed by the custom gripper during the rubbing motion was also tuned to enhance model interpretability. With the DenseTact camera frame rate of 10 Hz, a period of 4 seconds was found to be fast enough to produce noticeable changes in optical flow without aliasing between frames.

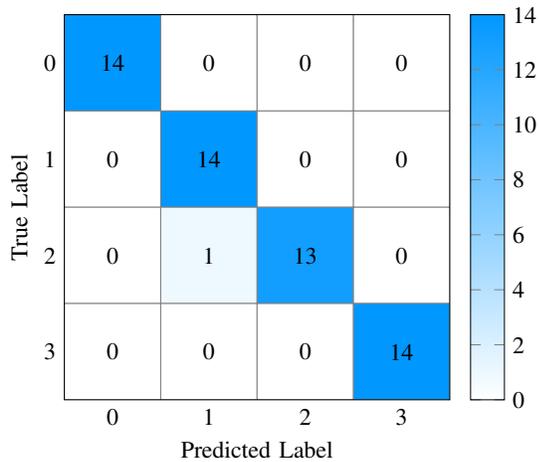
\begin{figure}[ht]
    \centering
    \scalebox{0.9}{\begin{tikzpicture}
    \begin{axis}[
            colormap={bluewhite}{color=(white) rgb255=(0, 151, 255)},
            xlabel=Predicted Label,
            ylabel=True Label,
            xticklabels={0, 1, 2, 3},
            xtick={0,...,3},
            xtick style={draw=none},
            yticklabels={0, 1, 2, 3},
            ytick={0,...,3},
            ytick style={draw=none},
            enlargelimits=false,
            colorbar,
            colorbar style={
                ytick={0, 2, 4, 6, 8, 10, 12, 14},
                point meta min=0,
                point meta max=14,
            },
            axis equal image, 
            nodes near coords={\pgfmathprintnumber\pgfplotspointmeta},
            nodes near coords style={
                yshift=-7pt
            },
        ]
        \addplot[
            matrix plot,
            mesh/cols=4,
            point meta=explicit,draw=gray
        ] table [meta=C] {
            x y C
            0 0 14
            1 0 0
            2 0 0
            3 0 0
            
            0 1 0
            1 1 14
            2 1 0
            3 1 0
            
            0 2 0
            1 2 1
            2 2 13
            3 2 0
    
            0 3 0
            1 3 0
            2 3 0
            3 3 14
            
        };
    \end{axis}
\end{tikzpicture}}
    \caption{\textbf{Cloth Layer Confusion Matrix:} Results of the classification for 56 test trials with the best-performing model. Only one trial from unseen test data was misclassified.}
    \label{fig:confusion_matrix}
\end{figure}

\begin{figure}[ht]
    \centering
    \scalebox{0.9}{\begin{tikzpicture}
    \begin{axis}[
        width=10cm, height=7cm,
        xlabel={T-SNE 1},
        ylabel={T-SNE 2},
        legend pos=south east,
        grid=major,
        legend style={font=\scriptsize},
    ]

    \addplot[
        only marks,
        mark=*, mark size=1.5pt,
        opacity=0.5,
        color=blue,
    ]
    coordinates {
        (18.9842, 3.4807444)
        (15.580391, 2.7526672)
        (16.275982, 2.9271708)
        (15.885907, 2.6885695)
        (15.947137, 2.7862277)
        (17.182753, 2.693505)
        (16.653387, 2.8791533)
        (15.436702, 2.6643963)
        (15.4587, 2.49797)
        (15.808395,	2.8246884)
        (15.251192, 2.6844003)
        (17.004887, 2.895945)
        (15.554294, 2.59753)
        (14.96655, 2.0938187)
    };

    \addplot[
        only marks,
        mark=*, mark size=1.5pt,
        opacity=0.5,
        color=orange,
    ]
    coordinates {
        (18.073637, -0.121873535)
        (18.283339, 0.12357941)
        (17.135506, 0.39128712)
        (17.182764, 0.4017135)
        (17.217289,	0.24785209)
        (18.404943,	0.42711562)
        (17.78871,	0.58790636)
        (17.697033,	0.37831125)
        (18.281328,	0.48826206)
        (17.405428,	0.42116264)
        (17.702984,	0.23860213)
        (18.03702,	0.9649428)
        (17.999552,	0.8358782)
        (19.018457,	1.288567)
    };
  
    \addplot[
        only marks,
        mark=*, mark size=1.5pt,
        opacity=0.5,
        color=green,
    ]
    coordinates {
        (16.507385,	0.33252463)
        (15.800205,	0.37074798)
        (16.037323,	0.19149156)
        (16.355059,	0.07819331)
        (15.76519,	0.3542355)
        (16.597046,	0.37071484)
        (15.962827,	0.35089052)
        (16.16801,	0.25938904)
        (15.667292,	0.5084365)
        (17.415325,	1.4727482)
        (20.08008,	3.5579019)
        (19.523655,	3.6332533)
        (20.443682,	3.3213472)
        (20.235094,	3.2191885)
    };
   
    \addplot[
        only marks,
        mark=*, mark size=1.5pt,
        opacity=0.5,
        color=red,
    ]
    coordinates {
        (19.959852,	3.0085895)
        (19.548843,	3.508198)
        (19.878029,	3.5400813)
        (19.518555,	2.9084127)
        (19.488813,	3.5660071)
        (19.271738,	2.9221644)
        (19.13973,	3.0755737)
        (19.068377,	3.2247772)
        (18.979376,	3.2796202)
        (17.780859,	3.0240698)
        (18.449839,	3.2122324)
        (19.41653,	2.764713)
        (19.733896,	2.8603384)
        (19.9616,	3.0569222)
    };

    \legend{Class 0, Class 1, Class 2, Class 3}
    \end{axis}
\end{tikzpicture}}
    \caption{\textbf{T-SNE Plot of Latent Features:} This plot visually represents high-dimensional data using a closeness metric. The latent feature space for the best-performing model shows the distribution of the feature vector across the classes for cloth classification. The separation of the classes allows the model to make accurate classifications of the cloth layers grasped. 3D T-SNE plots are available on the project website.}
    \label{fig:T-SNE}
    \vspace{-10pt}
\end{figure}
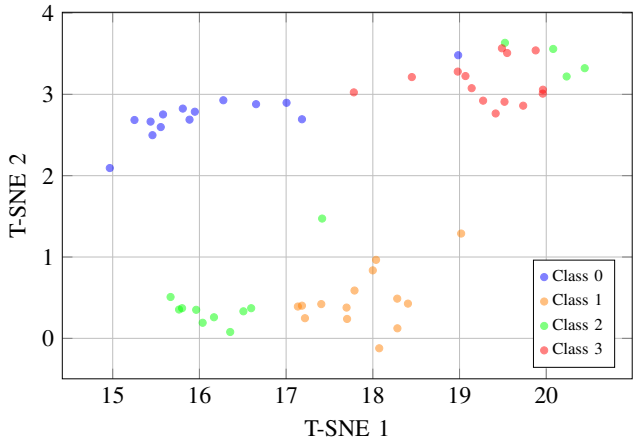

\subsection{Ablation Study} \label{section: ablation}
An ablation study was conducted to evaluate various approaches to the classification problem. Different network architectures and combinations of inputs were tested for their effect on model performance. Table \ref{tab:classifier_comparison} displays the results. 

\subsubsection{Input Ablations}
In addition to the full architecture in Fig. \ref{fig: architecture}, the effect of different inputs on performance was studied. As shown in Table \ref{tab:classifier_comparison}, the model was run using combinations of the following inputs: optical flow magnitudes and/or angles from one or both DenseTact sensors, 6-axis wrench, and joint angles. S1 was a calibrated DenseTact, providing both optical flow and wrench estimation, while S2 was an uncalibrated DenseTact, only providing optical flow data. Including all of the inputs in the CNN Transformer model resulted in the highest performance of 98.21\%. These results show the effectiveness of using the DenseTact's tactile information for this thin material layer classification task, achieving near-perfect classification on unseen test data.

\subsubsection{Architecture Ablations}
Different model architectures were also explored, modifying the complexity to accommodate different inputs. As shown in Table \ref{tab:classifier_comparison}, the ResNet-18 Transformer models performed similarly to the lower-performing CNN Transformer models. The lowest-performing model architecture was the naive CNN, which resulted in 25.00\% accuracy, equivalent to random guessing. Modifying this CNN to include batch norm layers and a dropout layer increased the test accuracy of the naive CNN. This modified CNN is denoted in Table \ref{tab:classifier_comparison} with an asterisk. Some transformer classifiers did not initially perform well (those with solely flow magnitudes as inputs), for which the architecture was also modified from a CNN to a CNN*. Using the transformer classifier with the CNN* backbone significantly increased test accuracy, due to the regularization in the CNN* and the transformer's attention mechanisms and ability to understand sequences of data.

\subsection{Cloth Classification Results}
As shown in Table \ref{tab:classifier_comparison}, the presented classifier model achieved a test accuracy of 98.21\% on cloth data. The test set of unseen cloth data consisted of 56 total trials (14 trials per class). Fig. \ref{fig:confusion_matrix} displays the confusion matrix from the cloth test trials. All test runs for 0, 1, and 3 layers of grasped cloth were correctly classified with 100\% accuracy, while only one trial from the 2-layer class was misclassified.

\subsection{Paper Classification Results}
In another set of trials, we perform the same experiments on layers of paper to evaluate our method's performance on other thin materials. Data was collected using the same process, with 50 trials each for 0, 1, 2, and 3 layers of paper. The top three performing model architectures for cloth were used, as noted in Table \ref{tab:classifier_comparison}, and trained using the data taken with paper layers. Table \ref{tab:model_comparison} shows the resulting test accuracies. While the models did not perform as well as they did with cloth, model 2 still achieved a test accuracy of 81.25\% for identifying the number of layers of paper. This discrepancy is likely due to more physical similarity between layered configurations of paper as compared to the thin cloth.

\subsection{Physical Interpretation} \label{physical}
As seen in the results from the ablation study, including optical flow magnitudes, net wrench, and joint state inputs increased accuracy on the test set. From a physics point of view, using all three inputs increased the model's ability to understand the friction forces at play during the rubbing motion where the DenseTacts and thin material interact. As seen in the videos of the rubbing motion available on the project website, the main noticeable difference between the classes is the amount of slipping, where a greater number of grasped layers increases the amount of slipping between the fingers since the surface of the materials can slide past each other more easily than the silicon surface of the DenseTacts. Although both the cloth and the paper were largely featureless visually, the dynamic tactile sensing was able to identify friction information and correctly classify layers, highlighting the usefulness of dynamic motions in robotic tasks. The reduced friction between paper layers relative to cloth layers may have caused more confusion between 2 layers and 3 layers of paper than between 2 layers and 3 layers of cloth. Additionally, the paper was less textured than the cloth, producing smaller deformations in the DenseTact sensors and reducing the optical flow information. Using optical flow, net wrench, and motor joint state data together gives the model more information to make accurate classifications of the layers grasped.

\section{Conclusion} \label{sec:conclusion}

In this work, we show the usefulness of high resolution, highly capable optical tactile sensors in combination with dynamic anthropomorphic motions to perform a difficult dexterous classification task. Specifically, we use DenseTact sensors mounted on a custom gripper to perform a rubbing motion, that is recorded and passed through a transformer-based neural network. This achieves near perfect accuracy on unseen data to classify the grasped layers of a thin silk cloth, and high accuracy on classifying the number of layers of paper. The gripper could be attached to a mobile robot, such as a LoCoBot as shown in Fig. \ref{fig:main}, to be used in real-world applications. This system can be incorporated into a two-stage pipeline, where a vision model first identifies the cloth type, and then this system accordingly grasps the object and performs layer classification with the corresponding network. This system can operate in real-time at 3 Hz to classify layers while the material is being rubbed. The cloth type and layer classification could then inform how the system manipulates the cloth for the relevant task, which could be bed-making, assistive dressing, or folding laundry, for example. 

While the results of this study are promising, there are many extension areas. Experimenting with different rubbing motions may improve feature differences. As depicted in the T-SNE \cite{TSNE} plot in Fig. \ref{fig:T-SNE}, the feature spaces for the classes have some overlap, and a different rubbing motion may help to further differentiate the feature spaces. Using a circular rubbing motion instead of the linear rubbing motion may provide more information about the non-linear friction forces between the sensors and the thin material. 

Exploring how this method extends to a larger output space, such as for classifying other thin materials and higher layer counts, is also potentially useful. We currently explore the capabilities for paper and cloth, but this method can be naturally extended to a variety of materials. In addition, this method can also be used to detect when the number of material layers grasped changes during a rubbing motion, for example, when manipulating a more crumpled material. Additionally, a future DenseTact sensor will enable roughness and  other more fine material attribute classification, expanding layer classification to more materials.

\addtolength{\textheight}{-5cm}   









  \bibliographystyle{./IEEEtran} 
  \bibliography{./IEEEexample}

\end{document}